# STACKING-BASED DEEP NEURAL NETWORK: DEEP ANALYTIC NETWORK ON CONVOLUTIONAL SPECTRAL HISTOGRAM FEATURES

*Cheng-Yaw Low, Andrew Beng-Jin Teoh*

School of Electrical and Electronic Engineering, Yonsei University, South Korea.

## ABSTRACT

Stacking-based deep neural network (S-DNN), in general, denotes a deep neural network (DNN) resemblance in terms of its very deep, feedforward network architecture. The typical S-DNN aggregates a variable number of individually learnable modules in series to assemble a DNN-alike alternative to the targeted object recognition tasks. This work likewise devises an S-DNN instantiation, dubbed deep analytic network (DAN), on top of the spectral histogram (SH) features. The DAN learning principle relies on ridge regression, and some key DNN constituents, specifically, rectified linear unit, fine-tuning, and normalization. The DAN aptitude is scrutinized on three repositories of varying domains, including FERET (faces), MNIST (handwritten digits), and CIFAR10 (natural objects). The empirical results unveil that DAN escalates the SH baseline performance over a sufficiently deep layer.

## 1. INTRODUCTION

Stacked generalization (SG) [1], in addition to bagging and boosting [2], was devised as a means of aggregating shallow learning models into a composition to improve the predictive force. The generic SG is of 2-level stacked-up architecture, where the level-1 input receives either the level-0 predictions (along with its true class labels), or the level-0 class probabilities, and the least-square regression is adopted as the level-1 generalizer for the classification tasks at hand. Some initial SG works are [1], [3], and [4].

The earliest stacking-based deep neural network (S-DNN), i.e., the cascade-correlation learning architecture [5], surfaced about the same time with SG. The latter S-DNNs, e.g., [6] and [7], primarily emphasize on the sequential labeling tasks. Other relatively modern S-DNNs are the deep neural network (DNN)-motivated exemplars, specifically, the deep belief network (DBN) [8], the deep Boltzmann machine (DBM) [9], and the deep autoencoder (DAE) [10], etc. One might assert that each of the restricted Boltzmann machine (RBM)-learned weight sets in DBN, as an example, would be globally fine-tuned via back-propagation like what DNNs do. The DBN network basis, however, still relies on the stack-up principle. Although a full understanding for S-DNN is somehow elusive (since there is still no theoretical foundations), the key intuition is that S-DNN decipher a large scale problem via modularization. Each S-DNN module, either be a sub-network, or merely an interleaved layer with non-linearity, learns an effective function individually to be stacked up in a chain, with the output of one feeding the input of the next. In other words, as there is no interaction between modules, each module is engaged to a pre-fixed problem, and learns to resolve that problem decisively and independently. Of course, the S-DNN learning speed would be much faster than DNN as it necessitates no back-propagation.

One of the inspiring exertions in the recent S-DNN literature is the deep convex network (DCN) [11]. Each DCN module indicates a specialized single hidden layer network yielding a ridge regression output set to the immediately adjacent module. The DCN innovation is extended to the kernel-DCN (K-DCN) [12], and K-DCN is further revised to tackle its scalability problem via random Fourier features for Gaussian kernel approximation [13]. The S-DNN closest to DCN are the deep extreme learning machine (D-ELM) variations, e.g., the stacked ELM (S-ELM) and the autoencoder-based stacked ELM (AE-S-ELM) [14], the hierarchical-ELM (H-ELM) [15], etc. If the input-hidden weights of DCN are randomly set, it can be generalized to the one equipping a series of ELMs. Other works delineated under this umbrella term are [16], [17], [18], etc.

This paper outlines a parsimonious S-DNN alternative to DCN, S-ELM, AE-S-ELM, and other counterparts, coined as deep analytic network (DAN) henceforth. The unique traits of DAN are: (i) DAN replaces the plurality of the layer-wise non-linear projections, either using random weights (as in S-ELM), or learnable weight sets, e.g., RBM-learned weights (as in DCN) and autoencoder-learned weights (as in AE-S-ELM), with the filter-based spectral histogram (SH) features [19]; (ii) the key DNN constituents, i.e., rectified linear unit (ReLU), proper fine-tuning, and normalization, are absorbed into the DAN pipeline; and (iii) our goal is not to compete with DNNs, but rather scrutinizing the extent to which the SH baseline performance can be advanced via the proposed stacking-based DAN. The 2-stage DAN schematic framework is portrayed in Fig. 1 as unsupervised, convolutional SH feature extraction in the first stage, followed by supervised, fully-connected DAN. Other details pertinent to DAN will be deliberated in the subsequent sections.

## 2. SPECTRAL HISTOGRAM FEATURES

The SH features are in line with DNN from the convolutional feature extraction perspective. The SH techniques, as a whole, pursue three stages: (i) a single-flat, or two-layer filter-image convolutions, either based on the learning-free Gabor filters, discrete Cosine transform (DCT) filters, etc., or the pre-learned filter ensembles, e.g., principal component analysis (PCA) filters, independent component analysis (ICA) filters, etc.; (ii) a non-linearity, i.e., binarization, followed by local binary pattern (LBP)-alike feature encoding; and (iii) a feature pooling stage via block-wise histogramming. The recent SH efforts include, but not limited to, binarized statistical image feature (BSIF) [20], PCA network (PCANet) [16], two-fold filter convolution (2-FFC) [21], etc. This work, however, only adopts BSIF, PCANet, and 2-FFC features as the DAN principal input to be post-appended with the layer-wise regression outputs.

## 3. DEEP ANALYTIC NETWORK

Similar to DCN and the D-ELM variants, DAN learns based on the long-existing ridge regression [22]. The primary reason leading us to equip DAN with RR is of its simplicity, and it offers an analytic solution. Suppose $\{(x_i, y_i)\}_{i=1}^N$ be $N$ BSIF, PCANet, or 2-FFC$_{PCA}$ features, where $x_i \in \mathcal{R}^d$, $y_i \in \{0,1\}^\mathcal{C}$ is the one-hot encoded vector indicating the class label for $x_i$, and $\mathcal{C}$ defines the number of training classes; the penalized residual sum of squares $\mathcal{E}$ is minimized as:

$$\mathcal{E}(W) = \mathrm{tr}\left[(Y - XW)^T(Y - XW)\right] + \lambda \|W\|_F^2 \quad (1)$$

where $X = [x_i, ..., x_N]^T \in \mathcal{R}^{N \times d}$, $Y = [y_i, ..., y_N]^T \in \mathcal{R}^{N \times \mathcal{C}}$, and $\|.\|_F$ denotes the Frobenius norm. Hence, $W \in \mathcal{R}^{d \times \mathcal{C}}$ is analytically estimated in batch mode as:

$$W = (X^T X + \lambda I)^{-1} X^T Y \quad (2)$$

where $I$ is an identity matrix of relevant dimension. For $N < d$, the primal solution in (2) can be expressed in its dual equivalence as:

$$W = X^T (X X^T + \lambda I)^{-1} Y \quad (3)$$

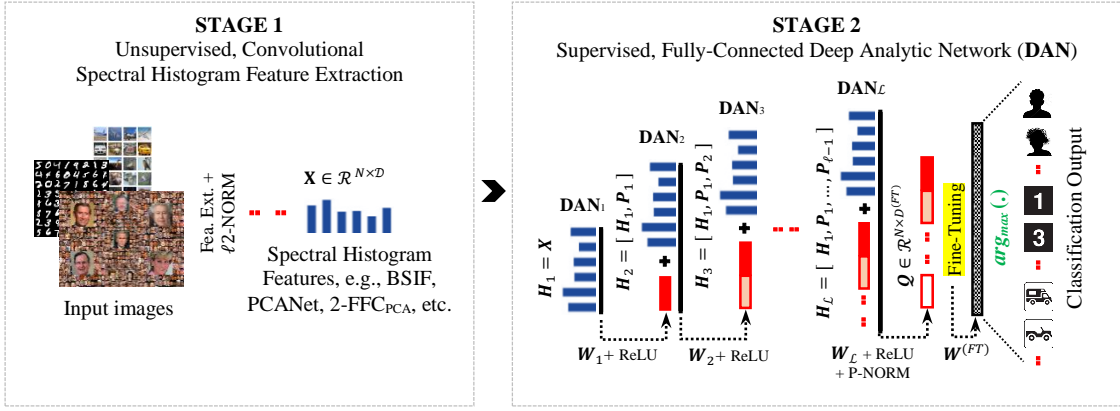

Fig. 1. Generic 2-stage DAN framework.

Note in our formulation that $X$ is to be $\ell2$-normalized by $N$; and $Y$ is of mean-centered. We omit the intercept term as $W$ is perceived to be biased owing to $Y$ centralization.

To begin with the $\mathcal{L}$-layer DAN learning (see Fig. 1), the first-layer DAN accepts $X$ and the targeted $W_\ell \in \mathcal{R}^{D_\ell \times \mathcal{C}}$, for $\ell = 1, ..., \mathcal{L}$, is estimated pursuant to (2) (in primal) as follows:

$$W_\ell = (H_\ell^T H_\ell + \lambda_\ell I)^{-1} H_\ell^T Y \quad (4)$$

where $H_1 = X$, and $H_{2,...,\mathcal{L}} = [H_1, P_1, ..., P_{\ell-1}]$, such that $H_\ell \in \mathcal{R}^{N \times D_\ell}$, and $D_\ell = d + \mathcal{C}(\ell - 1)$. As in (5), $P_\ell \in \mathcal{R}^{N \times \mathcal{C}}$ refers to the non-linearly transformed regression outputs of $H_\ell W_\ell$ with respect to ReLU.

$$P_\ell = \max(0, H_\ell W_\ell) \quad (5)$$

This implies that the negative regression outputs are simply regarded noises, and are therefore zeroed. In other words, we sparsify $H_\ell W_\ell$ to leverage DAN learnability before propagating it to the next layer to be a fragment of the effective input. We demonstrate in Section 4 that this non-negativity is of crucial.

In contrast to the modular fine-tuning (FT) practiced in DCN, the DAN FT layer, serving as the DAN output layer, synthesizes all $P_\ell$ into $Q$ to learn $W^{(FT)} \in \mathcal{R}^{D^{(FT)} \times \mathcal{C}}$ as follows:

$$W^{(FT)} = (Q^T Q + \lambda^{(FT)} I)^{-1} Q^T Y \quad (6)$$

providing that $Q = [P_1, ..., P_\mathcal{L}]^\beta \in \mathcal{R}^{N \times D^{(FT)}}$, $\beta$ is applied to $P_\ell$ in the point-wise manner, known as power-normalization (P-NORM), and $D^{(FT)} = \mathcal{L} \cdot \mathcal{C}$. In general, P-NORM regulates the disparities in $P_\ell$ before learning $W^{(FT)}$. We set $0.1 \leq \beta \leq 1$ in our experiments, and only the best one is reported. The RR-based FT layer, in fact, is parallel to the DNN softmax layer. Depending on the tasks at hand, it is opted for other classifiers, e.g., support vector machines (SVM).

To classify a query input, the pre-trained DAN is fed with the $\ell2$-normalized SH features $x \in \mathcal{R}^d$, and the conforming class label is predicted as:

$$cls(x) = \underset{k \in \{1,...,\mathcal{C}\}}{\mathrm{argmax}} \; q \, W^{(FT)} \quad (7)$$

where $q = [p_1, ..., p_\mathcal{L}]^\beta \in \mathcal{R}^{1 \times D^{(FT)}}$, and $p_\ell = \max(0, h_\ell W_\ell) \in \mathcal{R}^{1 \times \mathcal{C}}$, considering that $h_1 = x$, and $h_{2,...,\mathcal{L}} = [h_1, p_1, ..., p_{\ell-1}] \in \mathcal{R}^{1 \times D_\ell}$.

The entire DAN learning only involves three hyper-parameters: $\lambda_\ell$, $\lambda^{(FT)}$, and $\beta$. To be more precise, the DAN learning procedures are recapitulated in Table I. In a nutshell, DAN, DCN, and D-ELM share the RR-based learning principle. In addition to ReLU, and the supplementary FT output layer, the other distinguishable trait is that DAN works on the similar SH features in every layer, in conjunction to the regression outputs; whereas the sigmoidal projections in DCN and D-ELM depends on RBM-learned weights and random weights, respectively. Since the SH features are of high dimensional, we only compare DAN to K-DCN in Section 4.

## 4. EXPERIMENTAL RESULTS AND DISCUSSION

We access DAN with $\mathcal{L} = 10$ layers in the primal, or the dual form, depending on $N$, $D_\ell$, and $D^{(FT)}$. The DAN learning aptitude over the 10 layers is analyzed on three image classification tasks: (i) FERET [23] is a public face benchmarking dataset containing a gallery set, FA, and 4 probe sets with facial expression, illumination and time-span disturbances, i.e., FB, FC, DUP I, and DUP II. The FERET images are realigned with respect to the annotated eye coordinates and are each rescaled into 128×128 pixels; (ii) MNIST [24] consists of 70,000 gray-scale handwritten digits of each 28×28 pixels, where 60,000 images are for training and the rest for testing; (iii) CIFAR10 [25], on the other hand, possesses a sum of 60,000 natural images in RGB, with each 32×32 pixels. The training and testing sets are of 50,000 and 10,000 images, respectively. No other manipulations are applied, and our experiments also recruit no data augmentation. The BSIF, PCANet and 2-FFC features are abstracted beforehand based on the parameters configured in the respective papers, unless stated otherwise. The baseline performance for the original SH features is estimated using the naive nearest neighbor classifier with the Cosine similarity scores. For result replicability, the DAN hyper-parameters for each dataset is provided in Table II. These parameters are fine-tuned as a whole across $\mathcal{L}$ layers.

**Table I**. DAN step-by-step learning procedures.

| DAN LEARNING PROCEDURES |
|---|
| The DAN inputs are:<br>(i) $\ell 2$-normalized SH features $X \in \mathcal{R}^{N \times d}$;<br>(ii) Mean-centralized matrix of target labels $Y \in \mathcal{R}^{N \times \mathcal{C}}$;<br>(iii) DAN depth $\mathcal{L}$;<br>(iv) Layer-wise regularization parameters $\lambda_\ell$ and $\lambda^{(FT)}$;<br>(v) Power-normalization ratio $\beta$;<br>where $\mathcal{C}$ is the number of training classes, and $\ell = 1,2,\ldots,\mathcal{L}$. |
| Step 1: If $\ell = 1$, $H_1 = X$.<br>Otherwise, $H_\ell = [\,H_1, P_1, \ldots, P_{\ell-1}\,] \in \mathcal{R}^{N \times D_\ell}$,<br>where $D_\ell = d + \mathcal{C}(\ell - 1)$, and $\ell = 2,\ldots,\mathcal{L}$. |
| Step 2: Compute $W_\ell \in \mathcal{R}^{D_\ell \times \mathcal{C}}$ with respect to (4). |
| Step 3: Compute $P_\ell = \max(0, H_\ell W_\ell)$,<br>where $P_\ell = \in \mathcal{R}^{N \times \mathcal{C}}$. |
| Step 4: If $\ell \neq \mathcal{L}$, repeat Step 1 to Step 3 until $\ell = \mathcal{L}$.<br>Otherwise, $Q = [\,P_1, \ldots, P_\mathcal{L}\,]^\beta$,<br>where $Q \in \mathcal{R}^{N \times D^{(FT)}}$, and $D^{(FT)} = \mathcal{L} \cdot \mathcal{C}$. |
| Step 5: Compute $W^{(FT)} \in \mathcal{R}^{D^{(FT)} \times \mathcal{C}}$ with respect to (6). |

**Table II**. DAN hyper-parameter configuration with respect to varying datasets.

| PARAMETER | FERET | MNIST | CIFAR10 |
|---|---|---|---|
| $\lambda_{1,\ldots,\mathcal{L}}$ | 0.1 | 10 | 10 |
| $\lambda^{(FT)}_{1,\ldots,\mathcal{L}}$ | 0.1 | 10 | 10 |
| $\beta$ | 0.5 | 0.6 | 0.6 |

### 4.1. Performance Analysis

The DAN pipeline is decomposed into 4 instances to assess its layer-wise learning aptitude benefitted from the deep stacked-up building blocks, and the absorbed DNN essentials, i.e., ReLU and FT: (i) with no ReLU and no FT (linear DAN); (ii) with FT only (linear DAN with FT); (iii) with ReLU only; and (iv) with ReLU and FT. We observe from Table III that, for the 10-layer DAN with the original 2-FFC$_{PCA}$ features, the linear configuration fails to trigger DAN to learn after first layer - the rank-1 recognition rate is capped at 94% on the FERET DUP II probe set. If the linear DAN is affixed with a RR-based FT layer, its performance is progressed from the minima of 44.02% to 95.30%. To recap, the FT layer only accepts the power-normalized regression outputs $Q$ as the inputs, without the 2-FFC$_{PCA}$ features (refer to Section 3). In other words, for the single-flat layer DAN, the FT layer is forced to learn from the injudicious $P_1$ leading to a drastic performance drop. For the DAN with a sufficient depth, the $\mathcal{L}$ stacked-up regression outputs, i.e., $P_1, \ldots, P_{10}$ in this case, are evidenced yielding meaningful features for the FT output layer. The DAN with only ReLU, like what the linear DAN experiences, shows no improvement after the first layer. This is due to the reason that $Q$ suffers from the disparity problem to be regularized via the power-normalization in the FT layer. We attest in our experiments that both ReLU and the FT output layer are of indispensable. The DAN with these two DNN absorptions escalates from the baseline of 83.76% to 97.01%.

**Table III**. Performance summary, in terms of rank-1 identification rate (%), for DAN of depth 10 using 2-FFC$_{PCA}$ on FERET DUP II probe set.

| DESCR. | DUP II | | | |
|---|---|---|---|---|
| 2-FFC$_{PCA}$ | 83.76 | | | |
| | LINEAR | W/ FT | W/ RELU ONLY | W/ RELU + W/ FT |
| 2-FFC$_{PCA}$ + DAN$_1$ | 94.02 | 44.02 | **94.02** | 95.73 |
| 2-FFC$_{PCA}$ + DAN$_2$ | **94.44** | 62.39 | 94.02 | 96.15 |
| 2-FFC$_{PCA}$ + DAN$_3$ | 94.44 | 79.06 | 94.02 | **97.01** |
| 2-FFC$_{PCA}$ + DAN$_4$ | 94.44 | 85.90 | 94.02 | 96.58 |
| 2-FFC$_{PCA}$ + DAN$_5$ | 94.44 | 90.60 | 94.02 | 96.15 |
| 2-FFC$_{PCA}$ + DAN$_{10}$ | 94.44 | **95.30** | 94.02 | 95.30 |

### 4.2. Performance Evaluation on FERET

To our knowledge, there is no S-DNN relevant works evaluating on FERET, except PCANet. We thus only compare the performance of DAN to that of K-DCN [12], where each K-DCN module is, in fact, a kernelized-ELM. Table IV reveals that, with DAN or K-DCN, the baseline performance is vastly intensified across the FERET probe sets, in particular the PCANet, and the 2-FFC$_{PCA}$ features. The DAN performance, on average, stands out over K-DCN. We discern that, for the relatively less discriminative features like BSIF, the K-DCN learnability comes to a halt immediately after the first layer. On the contrary, DAN continues learning as it grows deeper. We derive the BSIF features using only 8 natural image-learned ICA filters shared by the authors. We believe that the DAN performance in FERET is the best in the recent face recognition literature.

**Table IV**. Performance summary, in terms of rank-1 identification rate (%), for DAN and K-DCN based on the BSIF, PCANet, and 2-FFC$_{PCA}$ features on FERET. The DAN and K-DCN layer yielding the best performance is superscripted in parentheses.

| DESCR. | FB | FC | DUP I | DUP II | MEAN |
|---|---|---|---|---|---|
| BSIF [20] (ICPR, 2012) | 93.47 | 69.07 | 71.75 | 59.40 | 73.42 |
| BSIF + K-DCN | 99.25 (1) | 98.97 (1) | 87.67 (1) | 85.47 (1) | 92.84 |
| BSIF + DAN | 99.58 (6) | **100** (2) | 92.66 (7) | 89.74 (7) | 95.50 |
| PCANet [16] (TIP, 2015) | 95.56 | 99.48 | 86.29 | 84.62 | 91.49 |
| PCANet + K-DCN | 99.75 (2) | **100** (1) | 96.82 (2) | 95.30 (2) | 97.97 |
| PCANet + DAN | 99.75 (2) | **100** (1) | 97.92 (2) | 96.15 (2) | 98.46 |
| 2-FFC$_{PCA}$ [21] | 95.65 | 99.48 | 86.57 | 83.76 | 91.36 |
| 2-FFC$_{PCA}$ + K-DCN | 99.75 (1) | **100** (1) | 96.82 (2) | 95.30 (1) | 97.97 |
| 2-FFC$_{PCA}$ + DAN | **99.83** (2) | **100** (1) | 97.92 (3) | 97.01 (3) | **98.69** |

## 4.3. Performance Evaluation on MNIST

In addition to S-ELM and AE-S-ELM [14], the convolutional DBN (CDBN) [26], DBM [9], and the stacked denoising autoencoders (SDAE) [10] are compared in Table V. With the adoption of the SH features, DAN outperforms the remaining methods marginally. It is noteworthy that S-ELM and AE-S-ELM demands a deeper network to be outstanding in performance. It is remarked in the paper that S-ELM and AE are respectively built with 650 and 700 in depth; while the best DAN performance is achieved in the 7-th layer with 99.34% for the PCANet features, and the 8-th layer with 99.45% for the 2-FFC$_{PCA}$ features. In accordance with[1], the least generalization error on MNIST to date is of 0.21%, equivalent to an accuracy of 99.79%, achieved by the regularization of neural network (R-NN) [27]. We notice that R-NN augments the training images to learn 5 networks, where the concluding performance is decided via voting. However, its accuracy is voted to be 99.43%, similar to that of DAN, without data augmentation.

**Table V**. Performance summary, in terms of rank-1 identification rate (%), for DAN and other relevant counterparts on MNIST probe set, where * denotes the empirical results reported in the respective papers.

| DESCR. | MNIST | DESCR. | MNIST |
|---|---|---|---|
| PCANet | 98.81 | *S-ELM [14] (TC, 2015) | 98.50 |
| PCANet + DAN | 99.34 [(7)] | *AE-S-ELM [14] (TC, 2015) | 98.89 |
| 2-FFC$_{PCA}$ | 99.01 | *CDBN [26] (ICML, 2009) | 99.18 |
| 2-FFC$_{PCA}$ + DAN | **99.45** [(8)] | *DBM [9] (JMLR, 2010) | 99.05 |
| | | *SDAE [10] (JMLR, 2010) | 98.72 |
| | | *R-NN w/ data aug. [27] (ICML, 2013) | **99.79** |
| | | *R-NN w/o data aug. [27] (ICML, 2013) | 99.43 |

## 4.4. Performance Evaluation on CIFAR10

The CIFAR10 images are of relatively more challenging compared to those in FERET and MNIST (due to wide intra-class variations). The PCANet features are replicated with 40 spatial pyramid pooling (SPP) histogram features [28] in $4 \times 4$, $2 \times 2$ and $1 \times 1$ sub-regions. In the meantime, with the 2-FFC elasticity, we derive the composite SPP features, dubbed 2-FFC$_{COMP}$ in this section, with respect to the 2-FFC PCA filters, the 2-FFC ICA filters, and the 2-FFC PCA-ICA filters. To closely approximate the PCANet features, the 2-FFC$_{COMP}$ features are loaded with 39 SPP histograms, i.e., 13 for each 2-FFC offspring type, in the same sub-region grids.

In place of the RR-based FT, the linear SVM classifier with the penalty parameter $C = 10$ is plugged into the DAN output layer to adapt DAN in a generally more complicated task. Table VI discloses that the SVM-based DAN with the PCA-compressed (due to overly long features) PCANet and 2-FFC$_{COMP}$ features of 5000 dimensions improves the baseline performance by at least 13%. Besides that, we also examine DAN against feature fusion for additional performance gain. The concatenated PCANet and 2-FFC$_{COMP}$ features (refers to the row of A + B in Table VI) further rises the accuracy to 79.30% in the 3-th DAN layer.

For reference, the CIFAR10 official portal[2] underlines that the DNN standard performance with and without data augmentation are of 89% and 82%, respectively; the 3 top-ranked DNNs summarized in[1] achieves 96.53% for fractional max-pooling network (FMP-Net) [29], 95.59% for large all convolutional network (large ALL-CNN) [30], followed by 94.16% for layer-sequential unit-variance network (LSUV-Net) [31]. Note that, FMP-Net and LSUV-Net are renamed for self-explanatory convenience. The common grounds (in addition to the gradient descent-based training algorithms) are: (i) most of the DNNs learns a bag of networks; FMP-Net, as an example, involves 100 networks; (ii) aggressive data augmentation, as in FMP-NET, ALL-CNN, and LSUV-Net.

**Table VI**. Performance summary, in terms of rank-1 identification rate (%), for DAN and other relevant counterparts on CIFAR10 probe set, where * denotes the empirical results reported in the respective papers.

| DESCR. | CIFAR10 | DESCR. | CIFAR10 |
|---|---|---|---|
| PCANet + PCA 5000 | 63.52 | *D-ELM [17] (N. Comp.,2016) | 56% |
| A : PCANet + PCA 5000 + DAN$_{SVM}$ | 76.90 [(5)] | *DNN w/o data aug.[2] | 82% |
| 2-FFC$_{PCA}$ + PCA 5000 | 61.92 | *DNN w/ data aug.[2] | 89% |
| B : 2-FFC$_{PCA}$ + PCA 5000 + DAN$_{SVM}$ | 76.27 [(3)] | *FMP-Net [29] | **96.53** |
| A + B | 79.30 [(3)] | * Large ALL-CNN [30] (ICLR, 2015) | 95.59 |
| | | *LSUV-Net [31] (ICLR, 2016) | 94.16 |

## 5. CONCLUSION

This work outlines a deep analytic network (DAN), i.e., a stacking-based deep neural network (S-DNN) instantiation, which simulates the typical feedforward deep neural network (DNN) in terms of its deep architecture. One of the distinguishable traits over DNN is that, the outputs for each S-DNN learning module are stacked up to one another as it traverses deep into the network. Unlike other S-DNNs, either relying on layer-wise non-linear random projection, or other learning-based projection, DAN operates on the filter-based spectral histogram features. The empirical results disclose that the DAN with an ample depth leads to remarkable performance gain comparing to the respective baselines. To bridge the performance gap in between DAN and DNN, it will be revised to convey other DNN essentials, e.g., dropout, whitening normalization, global fine-tuning, to name just a few.

## 6. ACKNOWLEDGEMENT

This work was supported by the National Research Foundation of Korea (NRF) grant funded by the Korea government(MSIP) (NO. 2016R1A2B4011656).

---

[1] http://rodrigob.github.io/are_we_there_yet/build/

[2] https://www.cs.toronto.edu/~kriz/cifar.html

## 7. REFERENCES

[1] D. H. Wolpert, "Stacked generalization," *Neural Networks*, 5(2), pp. 241-259, 1992.

[2] L. Breiman, "Bias, variance and arcing classifiers," *Technical Report 460*, Department of Statistics, University of California, Berkeley, CA, 1996.

[3] L. Breiman, "Stacked regressions," *Machine Learning*, 24, pp. 49-64, 1996.

[4] K. M. Ting, and I. H. Witten, "Issues in stacked generalization", *J. Mach. Learn. Res.*, 10(1), pp. 271-289, 1999.

[5] S. E. Fahlman, and C. Lebiere, "The cascade-correlation learning architecture," in *Proc. NIPS*, pp. 524-532, 1990.

[6] W. W. Cohen, and V. R. d. Carvalho, "Stacked sequential learning," in *Proc. IJCAI*, pp. 671-676, 2005.

[7] D. Yu, S. Wang, and L. Deng, "Sequential labeling using deep structured conditional random fields," *IEEE J. Selected Topics in Signal Proc.*, 4(6), pp. 965-972, 2010.

[8] G. E. Hinton, S. Osindero, and Y. W. Teh, "A fast learning algorithm for deep belief nets," *Neural Computation*, 18(7), 1527-1554, 2006.

[9] R. Salakhutdinov, and H. Larochelle, "Efficient learning of deep Boltzmann machines," *J. Mach. Learn. Res.*, vol. 9, pp. 693-700, 2010.

[10] P. Vincent, H. Larochelle, I. Lajoie, Y Bengio, and P. A. Manzagol, "Stacked denoising autoencoders: Learning useful representations in a deep network with a local denoising criterion," *J. Mach. Learn. Res.*, 11, pp. 3371-3408, 2010.

[11] L. Deng, D. Yu, and J. Platt, "Scalable stacking and learning for building deep architectures," in *Proc. ICASSP*, pp. 2133-2136, 2012.

[12] L. Deng, G. Tur, X. He, and D. Hakkani-Tur, "Use of kernel deep convex networks and end-to-end learning for spoken language understanding," in *Proc. IEEE Workshop on Spoken Language Technology*, 2012.

[13] P. -S. Huang, L. Deng, M. Hasegawa-Johnson, and X. He, "Random features for kernel deep convex network," in *Proc. ICASSP*, pp. 3143-3147, 2013.

[14] H. Zhou, G.-B. Huang, Z. Lin, H. Wang, and Y. C. Soh, "Stacked extreme learning machines," *IEEE Trans. on Cybern.*, 45(9), pp. 2013-2025, 2015.

[15] W. Zhu, J. Miao, L. Qing, and G.-B. Huang, "Hierarchical extreme learning machine for unsupervised representation learning," in *Proc. IJCNN*, pp. 1-8, 2015.

[16] Chan, T. H., Jia, K., Gao, S., Lu, J., Zeng, Z. and Ma, Y., "PCANet: A simple deep learning baseline for image classification?" *IEEE Trans. Image Process.*, 24(12), 5017-5032, 2015.

[17] M. D. Tissera, and M. D. McDonnell, "Deep extreme learning machines: supervised autoencoding architecture for classification," *Neurocomputing*, 174A, pp. 42-49, Jan. 2016.

[18] Z. Lei, D. Yi, S. Z. Li, "Learning stacked image descriptor for face recognition", IEEE Trans. On Circuits Syst. Video Technol., 26(9), pp. 1685-1096, Sept. 2016

[19] Liu, X. and Wang, D., "Texture classification using spectral histograms," *IEEE Trans. Image Process.*, 12(6), 661-670, 2003.

[20] Kannala, J. and Rahtu, E., "BSIF: Binarized statistical image features," in *Proc. ICPR*, pp. 1363-1366, 2012.

[21] C. Y. Low, A. B. J. Teoh, and C. J. Ng, "Multi-fold Gabor, PCA and ICA filter convolution descriptor for face recognition," arXiv:1604.07057, 2016.

[22] A. E. Hoerl, and R. W. Kennard. "Ridge regression: Biased estimation for nonorthogonal problems," *Technometrics*, 12(3), pp. 55-67, 1970.

[23] Phillips, P. J., Moon, H., Rizvi, S. A. and Rauss, P. J., "The FERET evaluation methodology for face-recognition algorithms," *IEEE Trans. Pattern Anal. Mach. Intell.*, 22(10), 1090-1101, 2000.

[24] Y. Lecun, L. Bottou, Y. Bengio, and P. Haffner, "Gradient-based learning applied to document recognition," *Proceedings of the IEEE*, 86(11), pp. 2278-2324, 1998.

[25] A. Krizhevsky, "Learning multiple layers of features from tiny images," Technical Report, 2009.

[26] H. Lee, R. Grosse, R. Ranganath, and A. Ng, "Convolutional deep belief networks for scalable unsupervised learning of hierarchical representation," in *Proc. ICML*, pp. 609-616, 2009.

[27] L. Wan, M. Zeiler, S. Zhang, Y. LeCun, and R. Fergus, "Regularization of neural network using DropConnect," in *Proc. ICML*, 2013.

[28] K. He, X. Zhang, S. Ren, and J. Sun, "Spatial pyramid pooling in deep convolutional networks for visual recognition," in *Proc. ECCV*, 2014.

[29] B. Graham, "Fractional max-pooling," arXiv:1412.6071v4, 2015.

[30] J. T. Springenberg, A. Dosovitskiy, T. Brox, and M. Riedmiller, "Striving for simplicity: the all convolutional net," in *Proc. ICLR Workshop*, arXiv:1412.6806v3, 2015.

[31] D. Mishkin, and J. Matas, "All you need is a good init," in *Proc. ICLR*, arXiv:1511.06422v7, 2016.